\documentclass{article} 

\usepackage[USenglish]{babel}
\usepackage{graphicx}

\usepackage[usenames]{xcolor}
\definecolor{citecl}{rgb}{0, 0, .8} 
\definecolor{linkcl}{rgb}{.1, .6, .1} 

\definecolor{Blue}{rgb}{0,0,1}
\definecolor{Red}{rgb}{1,0,0}
\definecolor{Green}{rgb}{0,.8,0}
\definecolor{Magenta}{rgb}{1,0,1}

\definecolor{swap}{rgb}{0,0,1}
\definecolor{expand}{rgb}{0.44,0.44,0.44}
\definecolor{trws}{rgb}{0,1,1}
\definecolor{bp}{rgb}{1,0,0}
\definecolor{icm}{rgb}{0,1,0}
\definecolor{rlp}{rgb}{1,0,1}

\definecolor{coarse}{rgb}{0.753, 0.314, 0.302} 
\definecolor{fine}{rgb}{0.310, 0.506, 0.741}   
\definecolor{ours}{rgb}{.8, 0, 0} 

\usepackage{makeidx}
\usepackage{picins}
\usepackage{multirow}
\usepackage{amsfonts}
\usepackage{amssymb}
\usepackage{amsmath}
\usepackage{amsthm}
\usepackage{bbm}
\usepackage[bookmarks=true,bookmarksopen=true,bookmarksopenlevel=0,colorlinks,bookmarksnumbered=true,plainpages=false]{hyperref}
\usepackage[square,numbers]{natbib}
\usepackage[ruled,vlined]{algorithm2e} 
\usepackage[final]{pdfpages}

\usepackage{stmaryrd} 
\usepackage{floatflt}

\usepackage{slashbox} 

\usepackage{appendix} 

\newcommand{\comment}[1]{}

\newcommand{\deff}{\mbox{$\stackrel{\mbox{\tiny{def}}}{=}$}}

\newcommand{\VV}{\mbox{$\mathcal{V}$}}
\newcommand{\EE}{\mbox{$\mathcal{E}$}}

\pdfoutput=1

\usepackage{nips12submit_e,times}

\nipsfinaltrue

\title{A Multiscale Framework for Challenging Discrete Optimization}

\author{
Shai Bagon \qquad Meirav Galun\\
Department of Computer Science and Applied Mathematics\\
Weizmann Institute of Science\\
Rehovot, Israel \\
\texttt{www.wisdom.weizmann.ac.il/$\sim$\{bagon,meirav\}}\\
}

\begin{document}

\maketitle

\begin{abstract}
Current state-of-the-art discrete optimization methods struggle behind when it comes to challenging contrast-enhancing discrete energies (i.e., favoring different labels for neighboring variables).
This work suggests a multiscale approach for these challenging problems.
Deriving an algebraic representation allows us to coarsen any pair-wise energy using any interpolation in a principled algebraic manner.
Furthermore, we propose an energy-aware interpolation operator that efficiently exposes the multiscale landscape of the energy yielding an effective coarse-to-fine optimization scheme.
Results on challenging contrast-enhancing energies show significant improvement over state-of-the-art methods.
\end{abstract}

\section{Introduction}
\label{sec:intro}

We consider discrete pair-wise energies, defined over a (weighted) graph $\left(\VV, \EE\right)$:
\begin{eqnarray}
E\left(L\right)&=&\sum_{i\in\VV} \varphi_i\left(l_i\right) + \sum_{\left(i,j\right)\in\EE} w_{ij}\cdot \varphi\left(l_i,l_j\right) \label{eq:GenEng}
\end{eqnarray}
where $\VV$ is the set of variables and $\EE$ is the set of edges.
The sought solution is a discrete vector: $L\in\left\{1,\ldots,l\right\}^n$, with $n$ variables each taking one of $l$ possible labels, minimizing~(\ref{eq:GenEng}).

Most energy instances of form~(\ref{eq:GenEng}) considered in the literature are {\em smoothness preserving}: that is, assigning neighboring variables to the same label costs less energy.
Smoothness preserving energies include submodular \cite{Schlesinger2006}, metric and semi-metric \cite{Veksler2002} energies.
State-of-the-art optimization algorithms (e.g., TRW-S \cite{Kolmogorov2006}, large move \cite{Veksler2002} and dual decomposition (DD) \cite{Komodakis2011}) handle smoothness preserving energies well yielding close to optimal results.
However, when it comes to {\em contrast-enhancing} energies (i.e., favoring different labels for neighboring variables) existing algorithms provide poor approximations (see e.g., \cite[example 8.1]{Wainwright2005}, \cite[\S5.1]{Kolmogorov2006}).
For contrast-enhancing energies the relaxation of TRW and DD is no longer tight and therefore they converge to a far from optimal solution.

This work suggests a multiscale approach to the optimization of contrast-enhancing energies.
Coarse-to-fine exploration of the solution space allows us to effectively avoid getting stuck in local minima.
Our work makes two major contributions:
(i)~{\bf An algebraic representation} of the energy allows for a {\em principled} derivation of the coarse scale energy using any linear coarse-to-fine interpolation.
(ii)~{\bf An energy-aware} method for computing the interpolation operator which efficiently exposes the multiscale landscape of the energy.

Multiscale approaches for discrete optimization has been proposed in the past (e.g., \cite{Gidas1989,Perez1996,Felzenszwalb2006,Kohli2010,Komodakis2010,Kim2011}).
However, they focus mainly on accelerating the optimization process of smoothness preserving energies.
Furthermore, these methods are usually restricted to a diadic coarsening of grid-based energies, and suggest
``ad-hoc" and heuristic derivation of the coarse-scale energy (e.g., \cite[\S3]{Kohli2010}).
In contrast, our framework suggests a {\em principled} derivation of coarse scale energy using a novel energy-aware interpolation yielding low energy solutions.

\section{Multiscale Energy Pyramid}
\label{sec:unified}

Our algebraic representation requires the substitution of vector $L$ in~(\ref{eq:GenEng})
with an equivalent binary matrix representation $U\in\left\{0,1\right\}^{n\times l}$.
The rows of $U$ correspond to the variables, and the columns corresponds to labels:
$U_{i,\alpha}=1$ iff variable $i$ is labeled ``$\alpha$" ($l_i=\alpha$).
Expressing the energy (\ref{eq:GenEng}) using $U$ yields a quadratic representation:
\begin{eqnarray}
E\left(U\right)&=&Tr\left(DU^T+WUVU^T\right) \label{eq:EngU} \\
      & \mbox{s.t.} & U\in\left\{0,1\right\}^{n\times l},\ \sum_{\alpha=1}^l U_{i\alpha}=1 \label{eq:const-U}
\end{eqnarray}
where $W=\left\{w_{ij}\right\}$, $D\in\mathbb{R}^{n\times l}$ s.t. $D_{i,\alpha}\deff \varphi_i(\alpha)$, and $V\in\mathbb{R}^{l\times l}$ s.t. $V_{\alpha,\beta}\deff \varphi\left(\alpha,\beta\right)$, $\alpha,\beta\in\left\{1,\ldots,l\right\}$.
An energy over $n$ variables with $l$ labels is now parameterized by $\left(n, l, D, W, V\right)$.

Let $\left(n^f, l, D^f, W^f, V\right)$ be the fine scale energy.
We wish to generate a coarser representation $\left(n^c, l, D^c, W^c, V\right)$ with fewer variables $n^c<n^f$.
This representation approximates $E\left(U^f\right)$ using fewer {\em variables}: $U^c$ with only $n^c$ rows.

An interpolation matrix  $P\in\left[0,1\right]^{{n^f}\times{n^c}}$ s.t. $\sum_jP_{ij}=1$ $\forall i$, maps coarse assignment $U^c$ to fine assignment $PU^c$.
For any fine assignment that can be approximated by a coarse assignment $U^c$, i.e.,
$U^f = PU^c$, we can write eq.~(\ref{eq:EngU}):
\begin{eqnarray}
E\left(U^f\right)&=&Tr\left(D^f{U^f}^T+W^fU^fV{U^f}^T\right) = Tr\left(D^f{U^c}^TP^T+W^fPU^cV{U^c}^TP^T\right) \label{eq:EngC} \\
&=&Tr\Big(\underbrace{\left(P^TD^f\right)}_{\mbox{\normalsize $ \deff D^c$ }}{U^c}^T + \underbrace{\left(P^TW^fP\right)}_{\mbox{\normalsize $\deff W^c$}}U^cV{U^c}^T\Big) =  Tr\left(D^c{U^c}^T+W^cU^cV{U^c}^T\right) \nonumber \\ & = & E\left(U^c\right) \nonumber
\end{eqnarray}
We have generated a coarse energy $E\left(U^c\right)$
parameterized by $\left(n^c, l, D^c, W^c, V\right)$ that approximates the fine energy $E(U^f)$.
This coarse energy is {\em of the same form} as the original energy allowing us to apply the coarsening procedure recursively to construct an energy pyramid.

Our principled algebraic representation allows us to perform label coarsening in a similar manner.
Looking at a  different interpolation matrix $\hat{P}\in\left[0,1\right]^{\mbox{$l^f\times l^c$}}$,
we  interpolate a coarse solution by $U^{\hat{f}} \leftarrow U^{\hat{c}}\hat{P}^T$.
This time the interpolation matrix $\hat{P}$ acts on the {\em labels}, i.e., the {\em columns} of $U$.
The coarse labeling matrix $U^{\hat{c}}$ has the same number of rows (variables), but fewer columns (labels).
Coarsening the labels yields:
\begin{equation}
E\left(U^{\hat{c}}\right) = Tr\left( \left(D^{\hat{f}}\hat{P}\right)\mbox{$U^{\hat{c}}$}^T + WU^{\hat{c}} \left(\hat{P}^TV^{\hat{f}}\hat{P}\right)\mbox{$U^{\hat{c}}$}^T\right)
\label{eq:EngC-V}
\end{equation}
Again, we end up with the same type of energy, but this time it is defined over a smaller number of discrete labels:
$\left(n, l^c, D^{\hat{c}}, W, V^{\hat{c}}\right)$,
where $D^{\hat{c}} \deff D^{\hat{f}}\hat{P}$ and $V^{\hat{c}} \deff \hat{P}^T V^{\hat{f}} \hat{P}$.

Equations~(\ref{eq:EngC}) and~(\ref{eq:EngC-V}) encapsulate one of our key contributions:
Constructing an energy pyramid depends only on $P$.
For {\em any} interpolation $P$ it is straightforward to derive the coarse-scale energy in a {\em principled} manner.
But what is an appropriate interpolation?


\section{Energy-aware Interpolation}
\label{sec:matrix-P}

The effectiveness of the multiscale approximation of~(\ref{eq:EngC}) and~(\ref{eq:EngC-V}) heavily depends on the interpolation matrix $P$ ($\hat{P}$ resp.).
The matrix $P$ can be interpreted as an operator that aggregates fine-scale variables into coarse ones (Fig.~\ref{fig:multiscale}).
Aggregating fine variables $i$ and $j$ into a coarser one excludes from the search space all assignments for which $l_i\ne l_j$.
This aggregation is undesired if assigning $i$ and $j$ to different labels yields low energy.
However, when variables $i$ and $j$ are {\em in agreement } under the energy (i.e., assignments with $l_i=l_j$ yield low energy),
aggregating them together allows for efficient exploration of low energy assignments.
{\bf A desired interpolation aggregates $i$ and $j$ when $i$ and $j$ are in agreement under the energy}.

\begin{floatingfigure}{.5\linewidth}
\centering
\parpic[r][r]{\includegraphics[width=.25\linewidth]{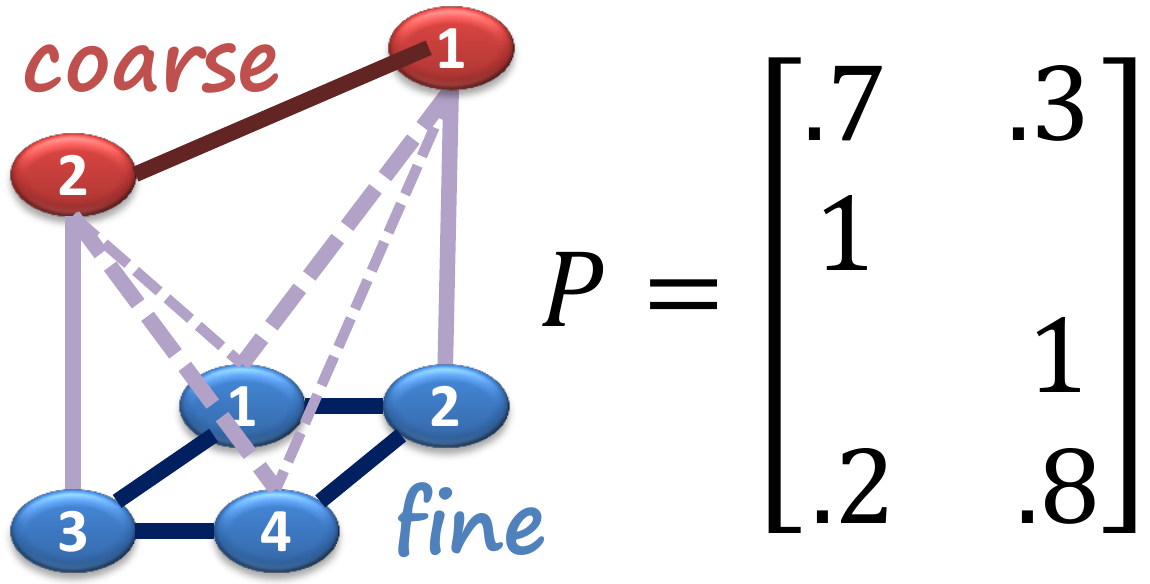}}
\caption{
{\bf Interpolation as soft variable aggregation:}
{\em
{\color{fine} fine} variables {\color{fine}1, 2, 3} and {\color{fine}4} are  softly aggregated into
{\color{coarse}coarse} variables {\color{coarse}1} and {\color{coarse}2}.
For example,  {\color{fine}fine} variable {\color{fine}1} is a convex combination of $.7$ of {\color{coarse}1} and $.3$ of {\color{coarse}2}.
Hard aggregation is a special case where $P$ is a binary matrix.
In that case each fine variable is influenced by exactly one coarse variable.}
}
\vspace*{2mm}
\label{fig:multiscale}
\end{floatingfigure}


To estimate these agreements we empirically generate several samples with relatively low energy,
and measure the label agreement between neighboring variables $i$ and $j$ in these samples.
We use Iterated Conditional Modes (ICM) \cite{Besag1986} to obtain locally low energy assignments. 
This procedure may be
interpreted as Gibbs sampling from the Gibbs distribution $p\left(U\right)\propto\exp\left(-\frac{1}{T}E\left(U\right)\right)$
at the limit $T\rightarrow 0$ (i.e., the ``zero-temperature" limit).
Performing $t=10$ ICM iterations with $K=10$ random restarts provides us with $K$ samples $\left\{L^k\right\}_{k=1}^K$.
The disagreement between neighboring variable $i$ and $j$ is estimated as $d_{ij}=\frac{1}{K}\sum_k V_{l^k_i,l^k_j}$, where $l^k_i$ is the label of variable $i$ in the $k^{th}$ sample.
Their agreement is then given by $c_{ij}=\exp\left(-\frac{d_{ij}}{\sigma}\right)$,
with $\sigma \propto \max V$.

Using the variable agreements, $c_{ij}$, we follow the Algebraic Multigrid (AMG) method of \cite{Brandt1986} to first determine the set of coarse scale variables and then construct an interpolation matrix $P$ that softly aggregates fine scale variables according to their agreement with the coarse ones.

We begin by selecting a set of coarse representative variables $\VV^c\subset \VV^f$,
such that every variable in $\VV^f \backslash \VV^c$ is in agreement with $\VV^c$.
A variable $i$ is considered in agreement with $\VV^c$ if $\sum_{j\in\VV^c}c_{ij} \ge \beta \sum_{j\in\VV^f} c_{ij}$.
That is, every variable in $\VV^f$ is either in $\VV^c$ or is {\em in agreement} with other variables in $\VV^c$,
and thus well represented in the coarse scale.

We perform this selection greedily and sequentially, starting with $\VV^c=\emptyset$ adding $i$ to $\VV^c$ if it is not yet in agreement with $\VV^c$.
The parameter $\beta$ affects the coarsening rate, i.e., the ratio $n^c/n^f$,
smaller $\beta$ results in a lower ratio.

At the end of this process we have a set of coarse representatives $\VV^c$.
The interpolation matrix $P$ is then defined by:
\begin{equation}
P_{iI(j)} = \left\{
\begin{array}{cl}
c_{ij}                & i\in\VV^f\backslash\VV^c,\ j\in\VV^c\\
1                      & i\in\VV^c, j=i\\
0                      & \mbox{otherwise}\\
\end{array}
\right. \label{eq:entries-of-P}
\end{equation}
Where $I(j)$ is the coarse index of the variable whose fine index is $j$ (in Fig.~\ref{fig:multiscale}: $I(2)=1$ and $I(3)=2$).

We further prune rows of $P$ leaving only $\delta$ maximal entries.
Each row is then normalized to sum to 1.
Throughout our experiments we use $\beta=0.2$ and $\delta=3$ for computing $P$.

\section{A Unified Discrete Multiscale Framework}
\label{sec:pipeline}

Given an energy $\left(n, l, D, W, V\right)$ at scale $s=0$,
our framework first works fine-to-coarse to compute interpolation matrices $\left\{P^s\right\}$ that construct the ``energy pyramid": $\left\{\left(n^s, l, D^s, W^s, V\right)\right\}_{s=0,\ldots,S}$.
Typically we reduce the number of variables by a factor of $2$ between consecutive levels, resulting with less than $10$ variables at the coarsest scale.
Since there are very few degrees of freedom at the coarsest scale ICM\footnote{Our framework is not restricted to ICM and may utilize other single-scale optimization algorithms.} is likely to obtain a low-energy coarse solution.
Then, at each scale $s$ the coarse solution $U^s$ is interpolated to a finer scale $s-1$: $\tilde{U}^{s-1} \leftarrow P^sU^s$.
At the finer scale $\tilde{U}^{s-1}$ serves as a good initialization for ICM (fractional solutions are rounded).
These two steps of interpolation followed by refinement are repeated for all scales from coarse to fine.

Our energy-aware interpolation and ICM play complementary roles in this multiscale framework.
ICM makes fine scale {\em local} refinements of a given labeling, while the energy-aware interpolation makes coarse grouping of variables to expose {\em global} behavior of the energy.
In a sense, ICM is a discrete equivalent to the continuous Gauss-Seidel relaxation used in continuous domain multiscale schemes.

\section{Experimental Results}
\label{sec:results}
We evaluated our multiscale framework on challenging contrast enhancing synthetic, as well as on co-clustering energies.
We follow the protocol of \cite{Szeliski2008} that uses the {\em lower bound} as a baseline for comparing performance of different optimization methods on different energies.
We report the ratio between the resulting energy and the lower bound
(in percents),
{\bf closer to $100\%$ is better}\footnote{Matlab implementation is available at: \url{www.wisdom.weizmann.ac.il/~bagon/matlab.html}}.

\begin{table}
\begin{minipage}[t]{.43\linewidth}
\caption{ {\bf Synthetic results:}
{\em
Showing percent of achieved energy value relative to the lower bound computed by TRW-S (closer to $100\%$ is better) for ICM and TRW-S
for varying strengths of the pair-wise term ($\lambda=5,10,15$, stronger $\rightarrow$ harder to optimize.)}
}
\centering
\begin{tabular}{c||c|c||c}
 \multirow{2}{*}{$\lambda$} & \multicolumn{2}{c||}{ICM} & \multirow{2}{*}{TRW-S} \\
 & {\color{ours}Ours}  & single scale &   \\\hline \hline
$5$ & {\color{ours}$112.6\%$} & $115.9\%$  & $116.6\%$ \\
$10$ & {\color{ours}$123.6\%$} & $130.2\%$ & $134.6\%$ \\
$15$ & {\color{ours}$127.1\%$} & $135.8\%$ & $138.3\%$ \\
\end{tabular}
\label{tab:res-synthetic}
\end{minipage}
\hspace*{.03\linewidth}
\begin{minipage}[t]{.52\linewidth}
\caption{ {\bf Co-clustering results: }
{\em
Baseline for comparison are state-of-the-art results of \protect\cite{Glasner2011}.
(a) We report our results as percent of the baseline: smaller is better, lower than $100\%$ even outperforms state-of-the-art.
(b) We also report the fraction of energies for which our multiscale framework outperform state-of-the-art.
} }
\centering
\begin{tabular}{c||c|c||c}
 & \multicolumn{2}{c||}{ICM} & TRW-S\\
               & {\color{ours}Ours}     & single scale & \\
\hline \hline
(a) & {\color{ours}$99.9\%$} & $177.7\%$ & $176.2\%$ \\\hline
(b) & {\color{ours}$55.6\%$} & $0.0\%$  & $0.5\%$ \\\hline
\end{tabular}
\label{tab:cocluster-res}
\end{minipage}
\end{table}

\noindent{\bf Synthetic:}~
We begin with synthetic {\em contrast-enhancing} energies defined over a 4-connected grid graph of size $50\times50$ ($n=2500$), and $l=5$ labels.
The unary term $D \sim \mathcal{N}\left(0,1\right)$.
The pair-wise term $V_{\alpha\beta}=V_{\beta\alpha} \sim \mathcal{U}\left(0, 1\right)$ ($V_{\alpha\alpha}=0$) and $w_{ij}=w_{ji} \sim \lambda \cdot \mathcal{U}\left(-1,1\right)$.
The parameter $\lambda$ controls the relative strength of the pair-wise term,
stronger (i.e., larger $\lambda$) results with energies more difficult to optimize (see \cite{Kolmogorov2006}).
The resulting synthetic energies are contrast-enhancing (since $w_{ij}$ may become negative).
Table~\ref{tab:res-synthetic} shows results, averaged over 100 experiments.
Using {\color{ours}our} multiscale framework to perform coarse-to-fine optimization of the energy yields significantly lower energies than single-scale methods used (ICM and TRW-S).


\noindent{\bf Co-clustering (Correlation-Clustering):}~
The problem of co-clustering addresses the matching of superpixels within and across frames in a video sequence.
Following \cite[\S6.2]{Bagon2012}, we treat co-clustering as a minimization of a discrete Potts energy adaptively adjusting the number of labels.
The resulting energies are contrast-enhancing (with some $w_{ij}<0$), have no underlying regular grid, no data term, and are very challenging to optimize.
We obtained 77 co-clustering energies, courtesy of \cite{Glasner2011}, used in their experiments.
Table~\ref{tab:cocluster-res} compares our discrete multiscale framework to the state-of-the-art results of \cite{Glasner2011} obtained by applying specially tailored convex relaxation method.
Our multiscale framework improves state-of-the-art for this family of challenging energies and significantly outperforms TRW-S.

\section{Extensions}
It is rather straightforward to extend our framework to handle energies with different $V$ for every pair $(i,j)$.
Moreover, higher order potentials can also be considered using the same algebraic representation.
A detailed derivation may be found in \cite{Bagon2012thesis}.

\subsubsection*{Acknowledgments}
We would like to thank Irad Yavneh, Maria Zontak and Daniel Glasner for their insightful remarks and discussions.
Special thanks go to Michal Irani for her exceptional encouragement and support.


\small{
\bibliographystyle{abbrv}
\bibliography{disc_opt,ms}
}


\end{document}